\documentclass[12pt,halfline,a4paper]{ouparticle}
\usepackage{amsmath}
\usepackage{csquotes}

\begin{document}

\title{Deep Learning Opacity in Scientific Discovery}

\author{%
\name{Eamon Duede}
\address{\small \textit{Department of Philosophy}}
\address{\small \textit{Committee on Conceptual \& Historical Studies of Science}}
\address{\small \textit{Pritzker School of Molecular Engineering}}
\address{\small \textit{Knowledge Lab}}
\address{\small University of Chicago}
\email{\small eduede@uchicago.edu}
}

\abstract{Philosophers have recently focused on critical, epistemological challenges that arise from the opacity of deep neural networks. One might conclude from this literature that doing good science with opaque models is exceptionally challenging, if not impossible. Yet, this is hard to square with the recent boom in optimism for AI in science alongside a flood of recent scientific breakthroughs driven by AI methods. In this paper, I argue that the disconnect between philosophical pessimism and scientific optimism is driven by a failure to examine how AI is actually used in science. I show that, in order to understand the epistemic justification for AI-powered breakthroughs, philosophers must examine the role played by deep learning as part of a wider process of discovery. The philosophical distinction between the `context of discovery' and the `context of justification' is helpful in this regard. I demonstrate the importance of attending to this distinction with two cases drawn from the scientific literature, and show that epistemic opacity need not diminish AI’s capacity to lead scientists to significant and justifiable breakthroughs.\\

\noindent **This paper is forthcoming in a supplementary issue of \textit{Philosophy of Science} containing selected
papers from PSA 2022. Final contents are subject to potential minor revision.**}

%Word Count: 4275}
\date{\today}

\maketitle
\section{Introductory}
\noindent The recent boom in optimism for the use of deep learning (DL) and artificial intelligence (AI) in science is due to the astonishing capacity of deep neural networks to facilitate discovery \cite{devries2018deep}, overcome the complexity of otherwise intractable scientific problems \cite{senior2020improved}, as well as to both emulate and outperform experts on routine \cite{chen2014deep}, complex, or even humanly impossible \cite{degrave2022magnetic} tasks. In fact, nearly every empirical discipline has already undergone some form of transformation as a result of developments in and implementation of deep learning
and artificial intelligence \cite{stevens2020ai}. To scientists and science funding agencies alike, artificial intelligence both promises and has already begun to revolutionize not only our science, but our society, and quality of life.\\

\noindent Yet, someone reading the recent philosophical literature on deep learning might be forgiven for concluding that doing good science with deep neural networks must be exceptionally challenging, if not impossible. This is because philosophers have, of late, focused not on the enormous potential of DL and AI, but on a number of important epistemological challenges that arise from the uninterpretability of deep neural networks (DNNs). Given that DNNs are epistemically opaque \cite{creel2020transparency,humphreys2009philosophical,zerilli2022explaining,lipton2018mythos}, it is, in many instances, impossible to know the high-level, logical rules that govern how the network relates inputs to outputs. It is argued that this lack of transparency severely limits scientists’ ability to form explanations for \cite{creel2020transparency,zerilli2022explaining} and understanding of \cite{sullivan2019understanding} why neural networks make the suggestions that they do. For instance, Creel states that ``access to only observable inputs and outputs of a completely opaque black-box system is not a sufficient basis for explanation[.]'' \cite[pg.573]{creel2020transparency}. So, this presents an obvious epistemic challenge when explanations of neural network logic are required to justify claims or decisions made on the basis of their outputs.\\ 

\noindent As a result, reading the recent philosophical literature can leave one wondering on what basis (beyond mere inductive considerations) neural network outputs can be justified \cite{boge2021two}. While lack of interpretability is of particular concern in high-stakes decision-making settings where accountability and value-alignment are salient (e.g., medical diagnosis and criminal justice) \cite{birch2022clinical, falco2021governing, hoffman2017taxonomy}, the opacity of deep learning models may also be of concern in basic research settings where explanations and understanding represent central epistemic virtues and often serve as justificatory credentials \cite{khalifa2017understanding}. Even if inductive considerations such as the past success of the model on out of sample data can help to raise confidence in the scientific merit of its outputs, in general, without additional justification, it is unclear how scientists can ensure that results are consistent with the epistemic norms of a given discipline.\footnote{To that end, scientists and philosophers have turned to various nascent approaches under the heading of Explainable AI (XAI) \cite{zerilli2022explaining,raz2022understanding}. Here, too, however, is the presumption that, without explanation of network logic, justification for network outputs will be hard to come by.} Or, so we are led to conclude.\\

\noindent There is, then, a sharp contrast between the relative optimism of scientists and policymakers on the one hand, and the pessimism of philosophers on the other, concerning the use of deep learning methods in science. This disconnect is due, I believe, to a failure on the part of philosophers to attend to the full range of ways that deep learning is actually used in science.\footnote{Though, I also believe that scientists routinely underestimate the epistemological challenge presented by DNN opacity.} In particular, while philosophers are right to examine and raise concern over epistemological issues that arise as a result of neural network opacity, it is equally important to step back and analyze whether these issues do, in fact, arise in practice and, if so, in what contexts and under what conditions.\\

\noindent In this paper, I argue that epistemological concerns due to neural network opacity will arise chiefly when network outputs are treated as scientific claims that stand in need of justification (e.g., treated as candidates for scientific knowledge, or treated as the basis for high-stakes decisions). It is reasonable to think that this must happen quite a bit, particularly outside of scientific settings. After all, the promise of deep learning is the rapid discovery of new knowledge. Of course, philosophers are correct that, if neural network outputs are evaluated in, what has often been referred to as, the ``context of justification'', then access to the high-level logic of the network (e.g., interpretability) will, in most cases, be required for validation.\footnote{This assumes, of course, that there are not compelling epistemological reasons to believe that neural network outputs are reliable that do not appeal to network logic, and do not rest on mere inductive grounds (e.g., model accuracy on holdout test sets). As of now, no such epistemology has been established.} While this certainly happens, I will show that scientists can make breakthrough discoveries and generate new knowledge utilizing fully opaque deep learning without raising any epistemological alarms. In fact, scientists are often well aware of the epistemological limitations and pitfalls that attend the use of black-box methods. But, rather than throw up their arms and embrace a form of pure instrumentalism (or worse, bad science), they can carefully position and constrain their use of deep learning outputs to what philosophers of science have called the ``context of discovery''.\\

\noindent The paper proceeds as follows: In \textit{Section 2}, I briefly describe neural networks and, drawing on recent philosophy of science \cite{creel2020transparency,zerilli2022explaining}, I explain the relevant sense in which deep learning models are opaque. In \textit{Section 3}, I sketch the epistemologically relevant distinction between treating neural network outputs as standing in need of justification (e.g., positioning outputs in the context of justification) and treating such outputs as \textit{part} of a wider process of conceiving new knowledge (e.g., positioning outputs in the context of discovery). In \textit{Section 4}, I present two cases which demonstrate the way in which researchers can make meaningful scientific discoveries by means of deep learning. Yet, in both cases, the findings themselves do not rely in any way on network interpretability nor accuracy for their justification.\\

\section{Deep Learning Opacity}

\noindent Deep learning is a machine learning technique based on artificial neural networks that is widely used for prediction and classification tasks \cite{lecun2015deep}. The goal of deep learning is to automate the search for a function $\hat{f}$ that approximates the true function $f$ that generates observed data. The fundamental assumption that motivates the use of deep learning is that $f$ is in the set of functions $\mathcal{F}$ \textit{representable} by a neural network given some particular architecture and (hyper)parameterization. Of course, for any given parameterization $k$, we have no way of knowing \textit{a priori} whether $f \in \mathcal{F}_k$. However, deep neural networks are universal approximators \cite{hornik1989multilayer}, so the assumption is at least principled.\\

% \noindent In a fully connected deep neural network architecture, every unit in every layer is connected to every unit in the preceding layer. We say that $a_n^l$ denotes the output of the $n^{th}$ unit in the $l_{th}$ layer, the value of which is obtained by applying an activation function $g$ to a weighted sum of its inputs from the previous layer such that $a_n^l = g(\sum_m w_{nm}^l a_m^{l-1})$. This can be rewritten to represent layer-wise output by vectorizing our components and keeping track of the weights at a given layer in a matrix $W\right$ which gives $\vartheta^l = g(W^l\right\vartheta^{l-1})$. In this way, mathematically, for any given input $x_i \in X$ we can represent the output at any layer $\vartheta_{l}(x)$ as

% \begin{equation}
% \vartheta(x_i)=\vartheta_{l}\left(\vartheta_{l-1}\left(\ldots \vartheta_{1}\left(x_i ; W_{1}\right) ; W_{2}\right) \ldots ; W_{l}\right)    
% \end{equation}

\noindent Like most regression tasks, the trained model $\hat{f}$ is arrived at by iteratively minimizing a loss function $\mathcal{L}$ through back-propagation of error gradients \cite{lecun2015deep} and updating all weights on all connections in the network such that the risk $R$ over the training distribution $P$ is minimized for $R_p(\hat{f}) := \mathbb{E}_{(\mathcal{X},\mathcal{Y})\sim P}[\mathcal{L}(\hat{f}(\mathcal{X}),\mathcal{Y})]$ where $\mathcal{X}$ and $\mathcal{Y}$ are sets of inputs and outputs. It is assumed that $\hat{f}$ is low risk over the distribution used for training if it performs well on a randomly selected iid test set from $P$. As a result, a highly accurate model is expected to perform well on out of training sample data which, in turn, provides a high degree of inductive support for confidence in the accuracy of its outputs. Nevertheless, while inductive considerations are common for assessing the merit of claims in science, they typically fall short of the justificatory standard of most disciplines.\\

\noindent From the above, it should be clear that there is a straightforwardly mathematical sense in which deep neural networks are \textit{fully transparent} \cite{leslie2019understanding,lipton2018mythos,zerilli2022explaining}. All weights on all connections across the network, billions as there may be, are both available to inspection and computationally tractable. However, while formally precise, neural network logic is largely semantically unintelligible. That is, the mathematical expression of a fully trained neural network model cannot, in general, be given an intelligible interpretation in terms of the target system such that one can understand or comprehend how the parts interact and contribute to the networks' outputs.\\

\noindent Zerilli \cite{zerilli2022explaining} describes the opacity of deep learning models (DLMs) by bringing out the distinction between ``Tractability'', ``Intelligibility'', and ``Fathomability'', a distinction echoed in \cite{lipton2018mythos}. Here, the idea is that any working machine learning model is tractable in so far as it can be run on a computer. However, intelligibility comes in degrees that are modulated by model fathomability. Fathomability is understood to be the extent to which a person can understand, straight away, how the model relates features to produce outputs. As a result, the more complex a model (e.g., increased dimensionality, extreme nonlinearities, etc.), the less fathomable it becomes. Many highly complex but linear models (e.g., random forests) remain ``intelligible'' in so far as all of the relationships between elements of the model can, in principle, be semantically deciphered even though the model as a whole (its overall decision logic) remains unfathomable due to complexity.\footnote{Consider a simple ordinary least squares (OLS) linear regression model: $y_i = \beta_0 + \beta_1 x_{i1} + \beta_2 x_{i2} + ... + \beta_k x_{ik} + \varepsilon_i$. Here, the variables have salient, semantic interpretations in terms of and corresponding to observed elements of the target. The model is at once intelligible and fathomable --the structure and composition of the model captures the relations of dependence (linear) between elements (semantics) in the model, and these can be read off directly.}\\

\noindent Zerilli's three aspects of epistemic access to neural network logic mirror Creel's three levels or granular scales of transparency  \cite{creel2020transparency}. For Creel, the transparency of a complex, computational model can be assessed ``Algorithmically'', ``Structurally'', and at ``Runtime''. Most relevant to the issues of this paper are algorithmic and structural transparency. For Creel, a model is algorithmically transparent if it is possible to establish which high-level, logical rules (e.g., which \textit{algorithm}) govern the transformation of input to output. In the case of a deep neural network, it is not possible to know which algorithm is implemented by the network precisely because the algorithm is developed autonomously during training. As a result, DNNs also lack what Creel calls ``structural'' transparency in that it is not clear how the distribution of weights and (hyper)parmeterization of the neural network implements (realizes) the algorithm that it has learned. Therefore, for Creel, DNNs are opaque ---neither ``fathomable'' nor ``intelligible'' in Zerilli's sense.\footnote{Both Creel and Zerilli draw on the computational concepts of understanding in information processing systems developed by David Marr \cite{marr2010vision,marr1977artificial}.} Following Humphreys \cite{humphreys2004extending,humphreys2009philosophical}, a process is said to be \textit{epistemically} opaque when it is impossible for a scientist to know all of the factors that are epistemically relevant to licensing claims on the basis of that process, where factors of `epistemic relevance' include those falling under Creel's algorithmic and structural levels and Zerilli's intelligibility and fathomability criteria. As such, DLMs are ``epistemically opaque''.

\section{Discovery and Justification with Deep Learning}

\noindent When it comes to justifying belief or trust in the outputs of deep learning models, their epistemic opacity is straightforwardly problematic. This is due to the fact that it is not possible to evaluate all of the epistemically relevant factors that led to the output. In high-stakes settings such as medical diagnosis, where the output of an epistemically opaque model forms the basis of a decision, a decision maker's inability to explain why the model prompts the decision that it does (and not, say, some other decision) can raise reasonable doubt as to whether the decision is, in fact, justified. Here is Creel on why we should strive for transparency:

\begin{displayquote}
``I claim that we should [strive for transparency] because scientists, modelers, and the public all require transparency and because it facilitates scientific explanation and artifact detection [...] Descriptively, the scientists who use the [epistemically opaque] systems to investigate, the modelers and computer scientists who create the systems, and the nonscientist citizens who interact with or are affected by the systems all need transparency.''\cite[pg.570]{creel2020transparency}
\end{displayquote}

\noindent Why might scientists (and others) all \textit{need} and \textit{require} transparency? The reasons Creel and most philosophers\footnote{There are, of course, exceptions including \cite{lenhard2010holism,humphreys2004extending,hooker2018machine}.} concerned with the epistemology of deep learning give are that, without transparency, scientists are unable to understand the outputs of their models, are powerless to explain why the models perform the way they do, cannot provide justification for the decisions they make on the basis of the model output, are uncertain whether and to what extent the models reflect our values ---on and on. What all of these reasons have in common is a commitment to the idea that neural network transparency is \textit{epistemically essential} to effectively use and gain knowledge from powerful artificial intelligence applications in scientific and societal settings \cite{gunning2019xai}.\\

\noindent However, I argue that there are many cases in which the epistemic opacity of deep learning models is \textit{epistemically irrelevant} to justifying claims arrived at with their aid. That is, it is justifiably possible to effectively use and gain scientific knowledge from epistemically opaque systems without sacrificing any justificatory rigor at all. In fact, scientists routinely achieve breakthroughs using deep learning that far exceed what they would have been able to do without such systems, while neither needing nor requiring transparency to justify their findings. This, I argue, can be readily observed when considering how epistemically opaque models can figure in the generation of findings.\\ 

\noindent Philosophers of science have, at least since Reichenbach \cite{reichenbach1938experience}, (and, later, Popper \cite{popper2002popper}) drawn a logical distinction between what has typically been referred to as the contexts of ``justification'' and ``discovery''. I argue that concerns about opacity are only epistemically relevant in settings where the outputs of epistemically opaque models are treated as candidates for scientific knowledge in their own right, that is, treated as claims that stand in need of justification. In these settings, neural network outputs are treated as the end result of an investigation (e.g., as findings in their own right) and, as such, fall within the ``context of justification''. However, in the context of justification, rigorous evaluation of the reasons that support findings is required to justify them. If those reasons are the internal logic of a neural network, then this kind of evaluation is blocked by neural network opacity.\footnote{A good, recent, example of this is the astounding breakthrough in deep learning enabled protein folding \cite{senior2020improved}}\\

\noindent Be that as it may, the outputs of epistemically opaque models need not be treated in this way. Rather, they can serve as aspects or parts of a \textit{process of discovery}. While the process ultimately leads to claims that stand in need of justification, the part played by an opaque model in that process can, itself, be epistemically insulated from the strong sort of evaluation that is applied to findings in the context of justification. In this way, neural network outputs can serve to facilitate discovery without their outputs or internal logic standing in need of justification. That is, neural network outputs that serve as parts of a process of discovery (similar to abduction \cite{duede2021social,hanson1965patterns} and problem-solving heuristics \cite{wimsatt2007re,simon1973does}) can be treated as situated in the ``context of discovery''.\footnote{Some have adopted a more nuanced view of the context of discovery as, itself, being split into a discovery part and a pursuit part. See, for example, \cite{laudan1981logic}.}\\

\begin{figure}[h]
    \centering
    \includegraphics[width=0.75\textwidth]{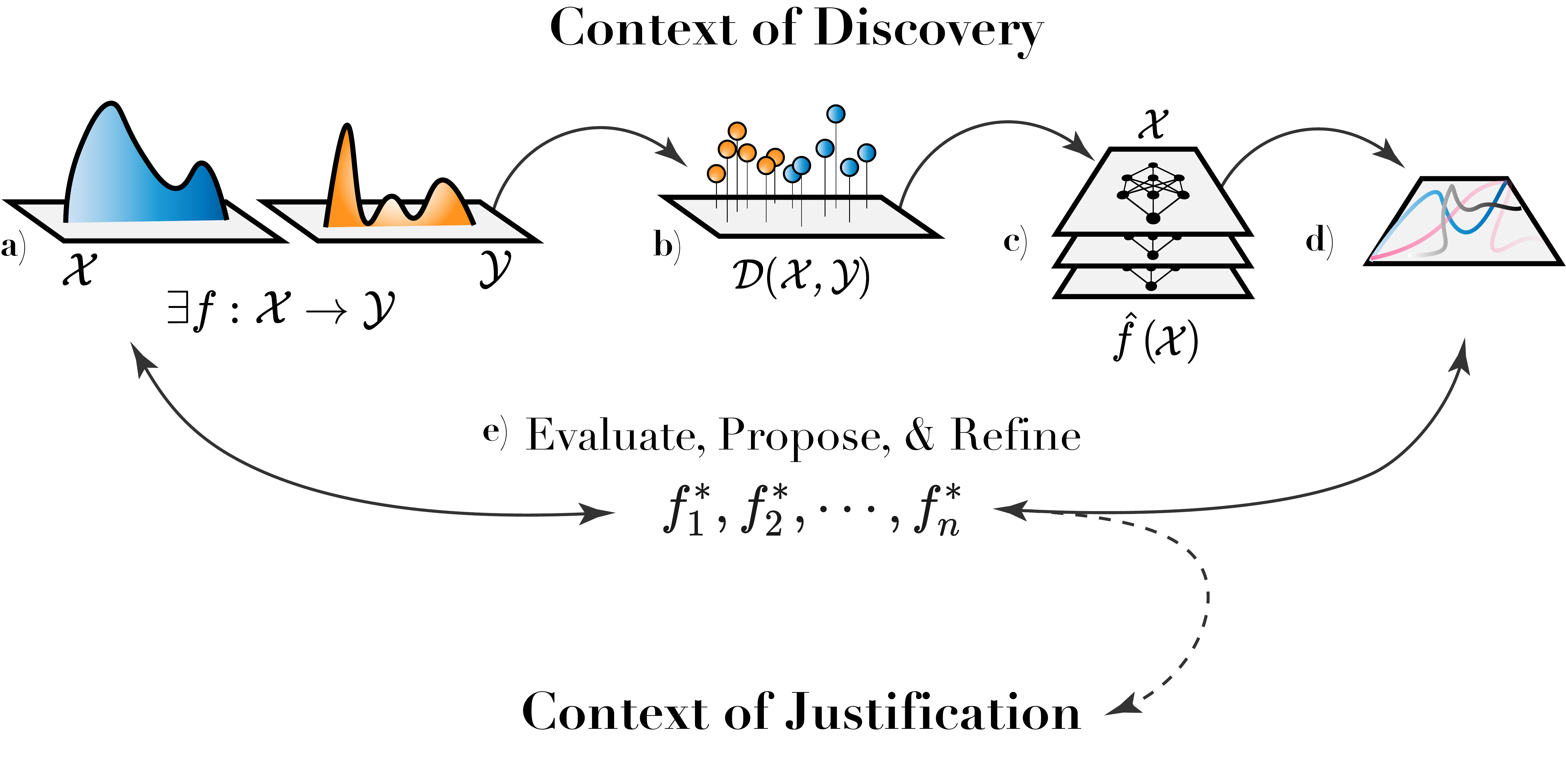}
    \caption{The confinement of epistemically opaque, neural network outputs to the context of discovery. \textbf{a)} posit or assume the existence of some theory $\exists{f}$ that connects two phenomena $\mathcal{X}$ and $\mathcal{Y}$; \textbf{b)} generate a dataset $\mathcal{D}$ that represents the assumed connection; \textbf{c)} train a deep learning model to learn a function $\hat{f}$ that approximates the posited theory; \textbf{d)} examine the behavior of $\hat{f}$; \textbf{e)} iteratively evaluate \textbf{(b)-(d)}, formulate, and refine conjectures or hypotheses $f_i^*$ that connect $\mathcal{X}$ and $\mathcal{Y}$. Justify $f^*$ by means distinct from those used to produce it.}
    \label{Figure 1}
\end{figure}
\noindent In the context of discovery, the outputs of neural networks can be used to guide attention and scientific intuition toward more promising hypotheses but do not, themselves, stand in need of justification. Here, outputs of opaque models serve to provide reasons to or evidence for pursuit of particular paths of inquiry over others (see: Figure 1). As such, they both provide and are subject to forms of preliminary appraisal \cite{schaffner1993discovery}, but, as the cases in \textit{Section 4} will bring out, the mere inductive support DLMs provide is epistemically sufficient to guide pursuit.\\

\section{Justified Discoveries Using Deep Learning}
\noindent In this section, I present two cases that demonstrate the claim that the epistemic opacity of DLMs can be epistemologically irrelevant for justifying the very scientific findings facilitated by their use.

\subsection{Case 1: Guiding Mathematical Intuition}
\noindent Here, I consider a case from low-dimensional topology in which researchers use deep learning to guide mathematical intuition concerning the relationship between two classes of properties of low-dimensional knots. Knots are particularly interesting topological objects because the relationships between their numerous properties are not well understood, and their various connections to other fields within mathematics are plausible but unproven.\\

\noindent In \cite{davies2021advancing}, mathematicians seek to discover and prove a conjecture that establishes a mathematical connection between known geometric and algebraic properties of knots in $\mathbb{R}^3$. In particular, the aim is to establish that hyperbolic invariants of knots (e.g., geometric properties of knots that are identical for all equivalent knots) and algebraic invariants of knots are connected. While the possibility of this connection had been imagined, its plausibility had not been established empirically and certainly not proved mathematically. As a result, mathematical intuition concerning a possible connection was too vague for genuine insight to emerge.\\

\noindent The deep learning facilitated approach in \cite{davies2021advancing} begins by \textit{imagining} that the geometric invariants $\mathcal{X}$ of a given knot $\mathcal{K}$ are, in fact, connected to that knot's algebraic invariants $\mathcal{Y}$. The algebraic invariant $\sigma \in \mathcal{Y}$ called the `signature' is known to represent relevant information about a given knot's topology. The researchers hypothesized that a knot's signature $\sigma(\mathcal{K})$ is provably related to its hyperbolic invariants $\mathcal{X(\mathcal{K})}$ in such a way that there exists some function $f$ such that $f(\mathcal{X(\mathcal{K})}) = \sigma(\mathcal{K})$ (Figure 1a). They then constructed a dataset of observed hyperbolic invariants and signatures for individual knots and trained a deep neural network to predict the latter from the former (Figure 1b). If the resulting DLM (Figure 1c) achieves an accuracy better than chance (here: $>0.25$) on a holdout set, then this provides researchers with reasons to expect that \textit{some} mathematical relationship must obtain between $\mathcal{X(K)}$ and $\sigma(\mathcal{K})$. Importantly, however, accepting this claim ultimately serves no role in justifying or proving the conjectured, mathematical relationship.\\

\noindent In fact, the initial DLM achieved an accuracy of roughly $0.78$, giving researchers high confidence in the belief that a connection between hyperbolic invariants and algebraic invariants does, in fact, obtain. While the established plausibility of the connection might be sufficient to find a promising conjecture, it is possible to isolate which of the various geometric invariants are most responsible for the accuracy of $\hat{f}$. Specifically, by quantifying the change in the gradient of the loss function with respect to each of the individual geometric invariants,\footnote{The saliency quantity $\mathbf{\mathrm{r}}$ can be calculated for each $\mathbf{x} \in \mathcal{X}$ by averaging the gradient of the cross-entropy loss function $\mathcal{L}$ with respect to each geometric invariant $x_i$ over all training examples such that $\mathrm{r}_{i}=\frac{1}{|\mathcal{X}|} \sum_{\mathbf{x} \in \mathcal{X}}\left|\frac{\partial L}{\partial \mathbf{x}_{i}}\right|$} it is possible to guide attention to a subset of hyperbolic invariants to consider when formulating a conjecture (Figure 1d). In particular, three such invariants are found to be most responsible for DLM accuracy in predicting signatures: the real and imaginary parts of the meridional translation $\mu$ and the longitudinal translation $\lambda$.\\

\noindent From these elements, mathematicians used their intuition to formulate an initial conjecture (Figure 1e) that relates $\mu(\mathcal{K})$ and $\lambda(\mathcal{K})$ to $\sigma(\mathcal{K})$ by means of a novel, conjectured property which they call the natural slope ($Re(\lambda / \mu)$) where $Re$ denotes the real part of the meridional translation) of $\mathcal{K}$. Using computational techniques that are common to experimental mathematics \cite{borwein2008mathematics}, corner cases were constructed that violate the initial conjecture, which was, in turn, refined into the following theorem.\\

Theorem: There exists a constant $c$ such that, for any hyperbolic knot $\mathcal{K}$,
\begin{equation}
    |2 \sigma(\mathcal{K})-\operatorname{slope}(\mathcal{K})| \leq c \operatorname{vol}(\mathcal{K}) \operatorname{inj}(\mathcal{K})^{-3}
\end{equation}

\noindent Let's call the above theorem $\mathcal{T}$. That $\mathcal{T}$ is true is provable. As a result, justification for belief in the truth of $\mathcal{T}$ does not lie in any empirical considerations. Nor does it rely on any facts about how the conjecture was arrived it. That is, the proof for $\mathcal{T}$ does not depend on any of the steps that were taken for its discovery. Its justificatory status is not diminished in any way by the fact that a number of assumptions were made in the process of its discovery nor that an opaque deep neural network aided in the decision to take seriously the connection between hyperbolic and algebraic properties.\\

\noindent One might object to the claim that the opacity of the network in this case was epistemically irrelevant. After all, the gradient based saliency method used to isolate the contribution to accuracy of the various inputs might be viewed as an interpretive step. While this objection is well taken, it is important to note that the saliency procedure used in this case was applied to the input layer which is, necessarily, transparent to begin with.\footnote{One knows and can interpret all inputs and their values.} Ultimately, then, what saliency methods add in this case is computational expediency as, alternatively, it was possible to use a simple combinatorial approach to iteratively search through the input space for the most predictive properties. In this way, saliency adds nothing of epistemic relevance to the process that would not have been possible without it. Moreover, the saliency is not required for justification. Nevertheless, as we will see, \textit{Case 2} is an example in which no such interpretive step is taken.

\subsection{Case 2: Deep Learning for Theory Improvement}

\noindent This case demonstrates the use of deep learning to dramatically improve our understanding of the geophysics of earthquakes by providing researchers good reasons to consider integrating known geophysical properties into existing theory. Consistent with the central claim of this paper, the neural network itself, while opaque, neither contributes nor withholds anything of epistemic importance to the justificatory credentials of the reworked theory.\\

\noindent The geophysics of earthquakes is poorly understood. In \cite{devries2018deep}, scientists seek to improve theory that describes the dynamics relating aftershocks to mainshocks. The best available theoretical models of aftershock triggering dynamics correctly predict the location of an aftershock with an $AUC = 0.583$. As the authors point out, while ``the maximum magnitude of aftershocks and their temporal decay are well described by empirical laws (such as Bath’s law and Omori’s law), [...] explaining and forecasting the spatial distribution of aftershocks is more difficult.'' \cite[pg.632]{devries2018deep}\\ 

\noindent To overcome this difficulty, scientists turn to deep learning to evaluate whether and to what extent it is possible to functionally relate mainshock and aftershock locations. After all, if an essentially stochastic process relates locations, then perhaps the extant theory is empirically adequate. As in \textit{Case 1}, researchers begin by imagining that aftershock locations are a function $f$ of mainshocks and seek to find an approximation of that function $\hat{f}$ (Figure 1a). To operationalize this, they construct a dataset of mainshock and aftershock events by representing the planet as a collection of 5km$^3$ tiles (Figure 1b). A tile is experiencing a mainshock, an aftershock, or no shock at any given moment. For the purposes of relating events, it is sufficient to treat the prediction as a simple, binary classification task (a task for which deep learning is particularly well suited) (Figure 1c). Given an input (mainshock parameter values and affected tiles), the task is to classify every terrestrial tile as either `aftershock' (1) or `not aftershock' (0).\\ 

\noindent The fully trained DLM correctly forecasts the locations of aftershocks between one second to one year following a mainshock event with an $AUC = 0.85$, significantly outperforming theory. This gives researchers good reason to take the possibility of further improving theory seriously (Figure 1d). Yet, this reason is not implicated in nor relevant to justifying the reworked theory.\\ 

\noindent The neural network outputs a probability distribution of `aftershock' over terrestrial tiles. The researchers compare this distribution to the one predicted by extant theory. Surprisingly, they observe that the probability of aftershock within a certain radius of a mainshock assigned by theory is largely uncorrelated with the probabilities assigned by the DLM. At this point, the predictive power of the DLM is treated as evidence that theory can be significantly improved thereby guiding the process of discovery.\\ 

\noindent Given the network's high accuracy, it is reasonable to assume that an improved theory would more closely resemble the observed probability distribution generated by the network. By examining the probability distributions forecast by the network and comparing them to the spatial distributions generated by known geophysical properties, scientists were able to narrow their search for parameters with with which to improve theory. By iteratively sweeping through known geophysical properties and correlating them with DLM distributions, they find that three parameters (maximum change in shear stress, the von Mises yield criterion, and aspects of the stress-change tensor), that had not been considered by geophysicists as relevant, in fact explain nearly all of the variance in predictions generated by the neural network, thereby providing novel physical insight into the geophysics of earthquakes (Figure 1e).\\

\noindent In this case, a fully opaque DLM has had profound implications for our theoretical understanding of earthquake dynamics. Namely, the ability to accurately predict phenomena orients scientific attention to empirical desiderata necessary for more accurate theory building. Moreover, it is epistemically irrelevant to justifying the improved theory that we cannot verify whether and how any of the geophysical quantities that were determined to be of relevance are, in fact, represented in the network. This is because it is not the network's predictions that stand in need of justification but, rather, the theory's itself. The reworked theory is justified in ways that are consistent with the norms of the discipline ---it relates known geophysical properties in ways that are consistent with first principles, it aids in the explanation and understanding of aftershock dynamics, and it outperforms extant theory in prediction. Yet, none of this depends on the neural network that was used to lead attention to relevant revisions of the theory for justification.\footnote{It is interesting to note that none of the relevant geophysical properties in \textit{Case 2} are in the input or output space of the neural network. So, there is no reason to think that these properties are modeled explicitly by the network. As a result, we have what Wimsatt might call a false model that, nevertheless, leads to, as he puts it, ``truer theory'' \cite{wimsatt1987false}.}

\section{Discussion}

\noindent What I hope to have shown in this paper is that, despite their epistemic opacity, deep learning models can be used quite effectively in science, not just for pragmatic ends but for genuine discovery and deeper theoretical understanding, as well. This can be accomplished when DLMs are used as \textit{guides} for exploring promising avenues of pursuit in the context of discovery. In science, we want to make the best conjectures and pose the best hypotheses that we can. The history of science is replete with efforts to develop processes for arriving at promising ideas. For instance, thought experiments are cognitive devices for hypothesis generation, exploration, and theory selection. In general, we want our processes of discovery to be as reliable or trustworthy as possible. But, here, inductive considerations are, perhaps, sufficient to establish reliability. After all, the processes by which we arrive at our conjectures and hypotheses do not typically serve also to justify them. While philosophers are right to raise epistemological concerns about neural network opacity, these problems primarily concern the treatment and use of deep learning outputs as findings in their own right that stand, as such, in need of justification which (as of now) only network transparency can provide. Yet, when DLMs serve the more modest (though no less impactful) role of guiding science in the context of discovery, their capacity to lead scientists to significant breakthroughs is in no way diminished.\\

\section{Acknowledgements}
Revised versions of this manuscript benefited greatly from conversations with Kevin Davey, James Evans, Ian Foster, Tyler Millhouse, Tom Pashby, Bill Wimsatt, and participants of the University of Texas at Austin's Philosophy of Biology Circle Working Group. This work was supported by the US \textit{National Science Foundation} \#2022023 NRT-HDR: AI-enabled Molecular Engineering of Materials and Systems (AIMEMS) for Sustainability.

\bibliographystyle{alpha}
\bibliography{main}

\end{document}